\documentclass[10pt,twocolumn,letterpaper]{article}

\usepackage{cvpr}
\usepackage{times}
\usepackage{epsfig}
\usepackage{graphicx, multirow}
\usepackage{amsmath}
\usepackage{amssymb}
\usepackage{makecell}
\usepackage{capt-of,etoolbox}

\usepackage[pagebackref=true,breaklinks=true,letterpaper=true,colorlinks,bookmarks=false]{hyperref}

\cvprfinalcopy % *** Uncomment this line for the final submission

\def\cvprPaperID{****} % *** Enter the CVPR Paper ID here

\ifcvprfinal\fi

\begin{document}

\title{Evading Deepfake-Image Detectors with White- and Black-Box Attacks}

\author{Nicholas Carlini \\
Google Brain \\
Mountain View, CA \\
{\tt\small ncarlini@google.com}
\and
Hany Farid\\
University of California, Berkeley\\
Berkeley, CA\\
{\tt\small hfarid@berkeley.edu}
}

%\cvprfinalcopy 

\maketitle

\begin{abstract}
It is now possible to synthesize highly realistic images of people who don't exist. Such content has, for example, been implicated in the creation of fraudulent social-media profiles responsible for dis-information campaigns. Significant efforts are, therefore, being deployed to detect synthetically-generated content. One popular forensic approach trains a neural network to distinguish real from synthetic content.

We show that such forensic classifiers are vulnerable to a range of attacks that reduce the classifier to near-$0\%$ accuracy. We develop five attack case studies on a state-of-the-art classifier that achieves an area under the ROC curve (AUC) of $0.95$ on almost all existing image generators, when only trained on one generator. With full access to the classifier, we can flip the lowest bit of each pixel in an image to reduce the classifier's AUC to $0.0005$; perturb $1\%$ of the image area to reduce the classifier's AUC to $0.08$; or add a single noise pattern in the synthesizer's latent space to reduce the classifier's AUC to $0.17$. We also develop a black-box attack that, with no access to the target classifier, reduces the AUC to $0.22$. These attacks reveal significant vulnerabilities of certain image-forensic classifiers.
\end{abstract}

% ------------------------------------------------------
\section{Introduction}
\label{sec:introduction}

\begin{figure}
    \centering
    \includegraphics[width=0.4\textwidth]{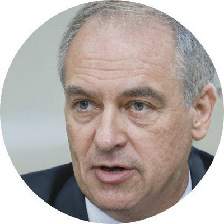}
    \bigskip
    
    \includegraphics[scale=0.25]{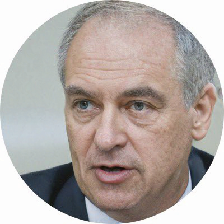} \raisebox{1cm}{$=$} \includegraphics[scale=0.25]{figures/Fig1/real.png} \raisebox{1cm}{$+$} \raisebox{1cm}{$\frac{1}{1000}$}\includegraphics[scale=0.25]{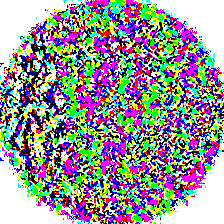}
    \\
    
    (a) \hspace{2cm} (b) \hspace{2.5cm} (c)
    \caption{Andrew Walz was, according to his Twitter account and webpage, running for a congressional seat in Rhode Island. In reality, Mr. Walz does not exist, and is the creation of a 17-year old high-school student. The profile picture (top) of the fictional candidate was synthesized using StyleGAN2~\cite{karras2019analyzing}. A state-of-the-art~\cite{wang20} synthetic-media detector would have flagged Mr. Walz's profile picture (b) as $87\%$ fake. We show, however, that adding a perceptually indistinguishable perturbation (c) to this photo causes the detector to classify the resulting picture (a) as $99\%$ real.}
    \label{fig:andrew-walz}
\end{figure}

According to his Twitter account, Andrew Walz, was a congressional candidate running for office in Rhode Island. He called himself ``a proven business leader'' with the tagline ``Let's make change in Washington together.'' Waltz's Twitter account was complete with his picture, Figure~\ref{fig:andrew-walz}, and a prized blue checkmark, showing that he had been \emph{verified} -- part of Twitter's efforts to verify the accounts of congressional and gubernatorial candidates.

Andrew Walz, however, was not real. He was the creation of a 17-year-old high-school student. During his holiday break, this student created a website and Twitter account for this fictional candidate~\cite{andrewwalz20}. The Twitter profile picture was plucked from the website \url{thispersondoesnotexist.com}. True to its name, and powered by StyleGAN2~\cite{karras2019analyzing}, this site generates images of people who don't exist.

The case of Mr. Walz's fictional congressional candidacy demonstrated how it might be possible to disrupt our democratic institutions through social-media powered dis-information campaigns. While this specific example was a fairly innocuous prank -- albeit exceedingly well executed -- recent reports have revealed how fake social-media accounts, with synthesized profile photographs, are being used by purported Russian hackers, trolls, and fraudsters~\cite{katiejones19,fakefaces20}. As dis-information campaigns continue to threaten our democratic institutions, civil society, and economic security, it has never been more important to be able to verify the contents of what we read, hear, and see on-line.

There are, therefore, significant efforts underway to develop forensic techniques to detect synthesized or manipulated audio, image, and video recordings. These  techniques can be partitioned into two broad categories: high-level and low-level. High-level forensic techniques focus on semantically meaningful features including, inconsistencies in eye blinks~\cite{li2018blinking}, head-pose~\cite{yang2019}, physiological signals~\cite{ciftci2019}, and distinct mannerisms~\cite{agarwal2019}.  Low-level forensic techniques detect pixel-level artifacts introduced by the synthesis process~\cite{yu18,marra18,rossler19,zhang19}. The benefit of low-level approaches is that they can detect artifacts that may not be visibly apparent. The drawback is that they, unlike high-level techniques, struggle to generalize to novel datasets~\cite{cozzolino18}, and can be sensitive to laundering (e.g.,~transcoding or resizing).

Recent work seemed to buck this trend of sensitivity and lack of generalizability~\cite{wang20,frank2020leveraging}. These techniques discriminate between real and synthetically-generated images that generalize across datasets and generators. In~\cite{wang20}, for example, the authors trained a standard image classifier on images synthesized by one technique (ProGAN~\cite{karras2017progressive}) and showed that this classifier detects synthesized images generated from nearly a dozen previously unseen architectures, datasets, and training methods. In addition, this classifier is robust to laundering through JPEG compression, spatial blurring, and resizing.

\smallskip
\noindent
\textbf{Contributions.} We find that neural networks designed to classify synthesized images~\cite{wang20,frank2020leveraging} are not \emph{adversarially robust}. Given an arbitrary image classified as fake, we can modify it imperceptibly to be classified as real. Building on work from the adversarial machine learning community~\cite{szegedy13,carlini2017towards,madry2018towards}, we investigate the robustness of forensic classifiers through a series of attacks in which it is assumed that we have (\emph{white-box}) or do not have (\emph{black-box}) full access to the classifier's parameters.

In line with prior work, we find that forensic classifiers are highly susceptible to such attacks. Our white-box attacks reduce the area under the ROC curve (AUC) from $0.95$ to below $0.1$ as compared to an AUC of $0.5$ for a classifier that randomly guesses ``real'' or ``fake''. Even when we are not able to directly access the classifier's parameters, our black-box attacks still reduce the ROC to below $0.22$.

% ------------------------------------------------------
\section{Background \& Related Work}

We begin by briefly reviewing techniques for creating and detecting synthetically-generated images as in Figure~\ref{fig:andrew-walz}.

\smallskip
\noindent
\textbf{Synthetically-Generated Images.} The most common approach to creating images of people (or cats, or objects) that don't exist leverages the power of generative adversarial networks (GAN). A GAN is composed of two main components, a generator and a discriminator. The generator's goal is to synthesize an image to be consistent with the distribution of a training dataset (e.g.,~images of people, cats, cars, or buildings, etc.). The discriminator's goal is to determine if the synthesized image can be detected as belonging to the training dataset or not. The generator and discriminator work iteratively, eventually leading the generator to learn to synthesize an image that fools the discriminator, yielding, for example, an image of a person who doesn't exist, Figure~\ref{fig:andrew-walz}. Following this general framework, dozens of techniques have emerged in recent years for synthesizing highly realistic content, including BigGAN~\cite{biggan18}, CycleGAN~\cite{cyclegan17}, GauGAN~\cite{gaugan19}, ProGAN~\cite{karras2017progressive}, StarGAN~\cite{stargan18}, StyleGAN~\cite{karras2018stylebased}, and StyleGAN2~\cite{karras2019analyzing}.

% ------------------------------------------------------

\smallskip
\noindent
\textbf{Detecting Synthetically-Generated Images.} Denote an image generator as $g \colon \mathcal{Z} \to \mathcal{X}$. The input to the generator is a vector in a latent space $\mathcal{Z}$, and the output is a color image of a pre-specified resolution. Denote an image-forensic classifier as $f \colon \mathcal{X} \to \mathbb{R}$. The input to the classifier is a color image, $x \in \mathcal{X}$, and the output is a real-valued scalar, where larger values correspond to a higher likelihood that the input image is fake or synthetically-generated.

We study the robustness of two classifiers: Wang \emph{et al.}~\cite{wang20} and Frank \emph{et al.}~\cite{frank2020leveraging}. The majority of our effort is focused on Wang \emph{et al.}, appearing jointly at CVPR'20 with this workshop, but consider Frank \emph{et al.} to show that our results are not limited to only one forensic classifier.

The forensic classifier of Wang \emph{et al.}~\cite{wang20} is based on ResNet-50~\cite{he16} pre-trained on ImageNet~\cite{deng09}, and then trained to classify an image as real or fake. The training dataset consists of a total of $720,000$ training and $4,000$ validation images, half of which are real images, and half of which are synthesized images created using ProGAN~\cite{karras2017progressive}. The images in this dataset are augmented by spatial blurring and JPEG compression. The accuracy of this classifier is evaluated against synthetically-generated images produced from ten different generators, similar in spirit, but distinct in implementation to the training images created by ProGAN. The trained classifier is not only able to accurately classify images synthesized by ProGAN, but also from ten other previously unseen generators. The classifier is also robust to simple laundering, consisting of spatial blurring and JPEG compression.

The forensic classifier of Frank \emph{et al.}~\cite{frank2020leveraging} takes a similar learning-based approach. The authors find that their classifier can accurately detect synthesized images from different generators. The authors argue that GAN synthesized-images have a common spatial frequency artifact that emerges from image upsampling that is part of the image-synthesis pipeline.

We will also consider a forensic classifier of our creation. This classifier is trained on $1,000,000$ ProGAN~\cite{karras2017progressive} images, half of which are real and half of which are fake. Our training pipeline is substantially simpler than~\cite{wang20}, and thus has an error rate that is roughly three times higher than~\cite{wang20}. The purpose of this classifier, however, is only to act as a mechanism for creating adversarial examples which can then be used to attack other classifiers.

\smallskip
\noindent
\textbf{Adversarial Machine Learning.} It is well established that machine learning classifiers are vulnerable to adversarial examples: for a classifier $f(\cdot)$ and input $x$, it is possible to construct an additive perturbation $\delta$ so that $x+\delta$ is misclassified by $f(\cdot)$~\cite{szegedy13}. Most strategies for constructing the perturbation $\delta$, rely on generating the perturbation with respect to the input $x$ through a gradient-descent optimization~\cite{carlini2017towards,madry2018towards}. While most adversarial machine learning work is on images, such attacks have also proven to be effective at attacking audio-~\cite{carlini18}, video-~\cite{neekhara20}, and text-~\cite{ebrahimi2017hotflip} classifiers. Orthogonal to defenses that detect synthesized content, related work has developed techniques to apply adversarial perturbations to unmodified images so that when a neural network attempts to modify it, the resulting image is meaningless~\cite{neekhara20}.

Attacking forensic classifiers should not simply be considered a sport.~\footnote{For this reason, we do not release our source code to (however marginally) increase the difficulty of repurposing our attacks.} Attacks and counter-measures are an important part of understanding the reliability of forensic techniques, and pushing these techniques to be more resilient to intentional and unintentional counter-measures~\cite{bohme13}. Indeed, concurrent work has addressed the strengthening of forensic classifiers~\cite{g2020adversarial,ruiz2020disrupting}.

% ------------------------------------------------------
\section{White-Box Attack}
\label{sec:white-box}

To begin, we evaluate the robustness of forensic classifiers to an attacker who has complete access to the classifier. This attacker is therefore able to compute the gradient of the input with respect to the classifier output, a so-called \emph{white-box} threat model. We apply three attacks that have previously been studied in the adversarial example literature, and then develop our own attack that subverts forensic classifiers by modifying the generator's latent space.

These attacks take the form of modifying synthesized images so that they are misclassified as real (with one exception where we devise an attack to misclassify a real image as fake). All images were obtained from the dataset of $94,036$ images released by~\cite{wang20}. On this dataset, their detector achieves an AUC~\footnote{AUC is defined as the area under the ROC curve that compares the false positive rate to the true positive rate. An AUC of $1.0$ corresponds to a perfect classifier, while an AUC of $0.5$ corresponds to a chance classifier.} of $0.97$. Even after performing typical laundering strategies in an attempt to conceal traces of synthesis (JPEG compression and spatial blurring), the classifier AUC remains above $0.94$ . 

In each attack described below, we operate on a randomly sampled subset of $10,000$ fake images. Each three-channel (RGB) image is of size $224 \times 224$ pixels with pixel intensities in the range $[0,1]$. The difference between two pixels will be measured in terms of a $0$-norm ($\ell_0$) or a $2$-norm ($\ell_2$). Flipping one pixel's RGB value, for example, from black $(0,0,0)$ to white $(1,1,1)$ yields an $\ell_0$ difference for this pixel of $3$ and an $\ell_2$ difference of $\sqrt{3}$.

\subsection{Distortion-minimizing Attack}
\label{sec:perturb}

Given a synthetically-generated image $x$ that is classified by $f(\cdot)$ as fake, we begin by constructing a small additive perturbation $\delta$ so that $x + \delta$ is instead incorrectly classified as real. A standard approach for computing an ``optimal'' perturbation $\delta$ relies on minimizing the $p$-norm $\lVert \delta \rVert_p$ for $p=\{0,1,2,\infty\}$~\cite{carlini2017towards}. Although the $p$-norm does not necessarily capture perceptual differences, for sufficiently small norms, such optimizations suffice to create impercetible image perturbations while revealing a classifier's fragility. Additionally, if attacks are possible under these $p$-norms, then attacks under less constrained norms are likely to be even more effective~\cite{carlini2017towards,gilmer2018motivating}.

While there are plethora of attacks, most follow a simple two-step process~\cite{madry2018towards}: (1) choose a loss function $L(x + \delta)$ so that $L(\cdot)$ is minimized when $f(x + \delta)$ is misclassified; and (2) minimize the loss function $L(\cdot)$ to obtain a perturbation $\delta$ that succeeds in decreasing classification accuracy. For the simple two-class problems (e.g.,~real or fake), where $f(x)$ is a scalar and our objective is to misclassify the image $x$ as real, it suffices to choose $L(x) = f(x)$.

In this setting we first describe an attack that directly minimizes the magnitude of the perturbation $\delta$ such that the resulting adversarial examples are classified as real. Let $\tau$ be a threshold such that when $f(x) < \tau$, an image is classified as real.\footnote{A drawback of this style of attack is that it requires a hard decision threshold $\tau$. In practice the value of this threshold depends on the acceptable false positive rate. We set $\tau=5\%$, a high value considering the low base rate of synthetic images in the wild.} The adversary then solves the following optimization problem:
\begin{eqnarray}
\mathop{\text{arg min }}_\delta \big( \lVert \delta \rVert_p \big), \quad \mbox{such that } f(x + \delta) < \tau.
%\mbox{minimize } \lVert \delta \rVert_p \quad \mbox{such that } f(x + \delta) < \tau.
\end{eqnarray}
This optimization formulation, however, is computationally intractable with standard gradient descent due to the nonlinear inequality constraint~\cite{szegedy13}. We, therefore, reformulate this optimization with a Lagrangian relaxation, which lends itself to a tractable gradient-descent optimization:
\begin{eqnarray}
  \mathop{\text{arg min }}_\delta \bigg( \lVert\delta \rVert_2 + c f(x+\delta) \bigg),
\label{eqn:mindistortion}  
\end{eqnarray}
where $c$ is a hyper-parameter that controls the trade-off between minimizing the norm of the perturbation $\delta$ with minimizing the loss $f(\cdot)$. A larger value of $c$ results in adversarial examples that are over-optimized (and more adversarial than they need to be), whereas a smaller value of $c$ results in a perturbation that is small -- as desired -- but not adversarial.

The optimization of Equation~(\ref{eqn:mindistortion}) proceeds as follows. For a given hyper-parameter $c$, the optimal $\delta$ is determined using gradient-descent minimization with the Adam optimizer~\cite{kingma2014adam} for $1,000$ iterations. An approximately optimal hyper-parameter $c$ is found through a binary search as follows. We initially consider values of $c_0=0$ and $c_1=100$ (or some sufficiently large value so that the attack is successful). The attack is then run with $c = {1 \over 2}(c_0+c_1)$. If the attack is successful, then $c_1=c$, otherwise $c_0=c$. This process is repeated until $c_0=c_1$.

This attack is effective but leads to such small distortions as to be impractical. In particular, saving the resulting adversarial image as an uncompressed PNG obliterates the attack because the image is quantized to $8$-bits per color channel.

We consider, therefore, a refinement to the $\ell_0$-distortion attack from~\cite{carlini2017towards} in which instead of minimizing the $\ell_2$ distortion, we minimize the fraction of pixels whose lowest-order bit needs to be flipped so that the image is misclassified. To do this, the above $\ell_2$ attack is applied with an additional constraint that the maximum perturbation to any pixel is $1/255$. After an adversarial image is generated, all pixels with the smallest perturbation are reset to their original value and these pixels are disallowed from future change. The attack then repeats, iteratively shrinking the set of perturbed pixels until convergence. With a maximum perturbation of $1/255$, this attack modifies a subset of pixels by, at most, flipping a pixel's lowest-order bit. In such an attack, the resulting adversarial image can be saved as an uncompressed PNG or even compressed JPEG image and still be misclassified. 

\paragraph{Attacking Wang \emph{et al.}~\cite{wang20}.} Directly applying this $\ell_2$-distortion minimizing attack is highly effective at attacking this forensic classifier. At a fixed false positive rate of $5\%$, an $\ell_2$-distortion of $0.02$ reduces the true positive rate to chance performance of $50\%$, while an $\ell_2$-distortion of $0.1$ reduces the true positive rate to just $0.1\%$.

Compared to the $\ell_2$-distortion of $0.02$ that reduces this forensic classifier to chance performance, reducing an ImageNet classifier (using the same model architecture on images of the same size) to chance performance requires a distortion over $16$ times larger~\cite{carlini2017towards}. These extremely small distortions suggest that the forensic classifier is highly sensitive and vulnerable to attack.

\begin{figure}
    \begin{center}
       %\begin{tabular}{c@{\hspace{0.1cm}}c}
        \raisebox{2.5cm}{(a)}
        \includegraphics[width=0.85\linewidth]{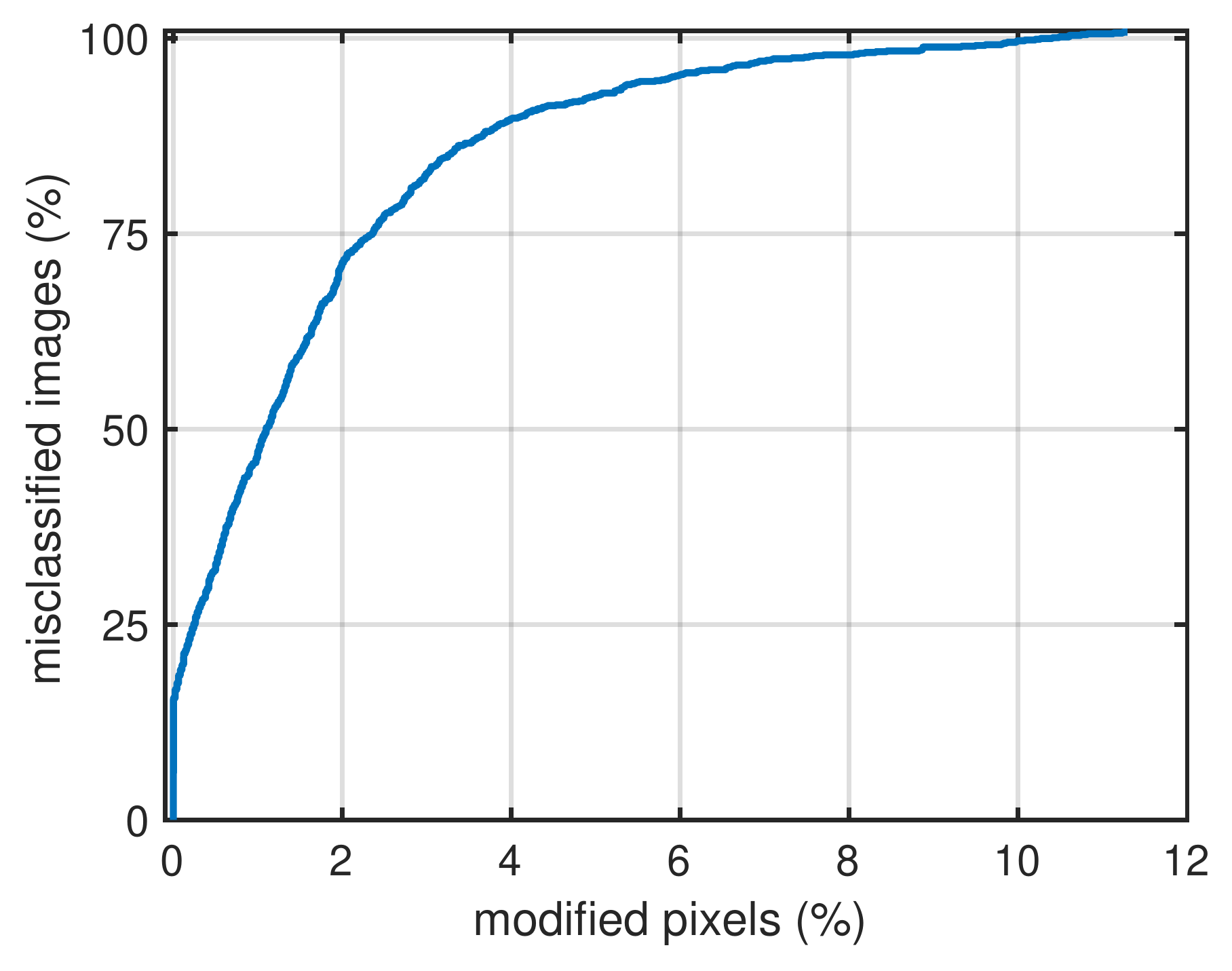} \\
        \raisebox{2.5cm}{(b)}
        \includegraphics[width=0.85\linewidth]{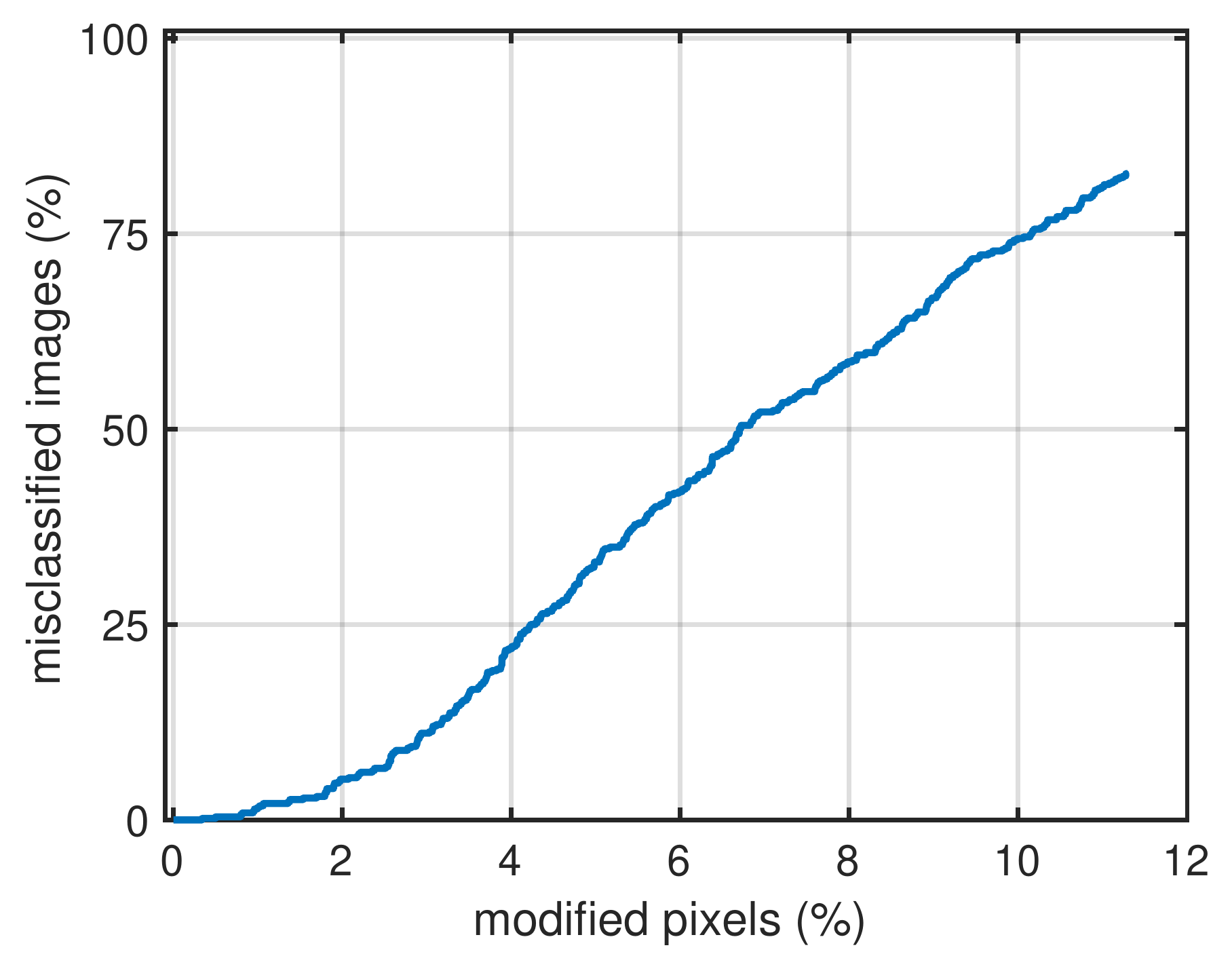} \\
       %\end{tabular}
    \end{center}
    \caption{The required $\ell_0$-distortion to fool the classifier into identifying (a) fake images as real or (b) real images as fake. Half of all fake images are misclassified as real by flipping the lowest-order bit of just $1\%$ of pixels. Half of all real images are misclassified as fake by flipping the lowest-order bit of less than $7\%$ of pixels.}
    \label{fig:lpattacks}
\end{figure}

The $\ell_0$-distortion minimizing attack is equally effective. Shown in Figure~\ref{fig:lpattacks}(a) is the percent of fake images misclassified as real as a function of the percent of modified pixels: with only $2\%$ pixel changes, $71.3\%$ of images are misclassified; with only $4\%$ pixel changes, $89.7\%$ of images are misclassified; and with less than $11\%$ pixel changes, nearly all images are misclassified.

\paragraph{Attacking Frank \emph{et al.} \cite{frank2020leveraging}.} After we developed the above attack, Frank \emph{et al.}~\cite{frank2020leveraging} released their study and corresponding pre-trained classifiers. A similar attack was applied to this classifier. This classifier not only detects if an image is synthetically-generated, but also predicts the identity of the generator. We therefore slightly modified our attack: instead of reporting success on any misclassification (e.g.,~reporting that a ProGAN image was generated by BigGAN), we only report success if the image is classified as real. Despite this increased discriminative performance, we find that we can reduce the true positive rate of the classifier on images generated by ProGAN from $99\%$ to $0\%$ by flipping the lowest-order bit of $50\%$ of the pixels.

\paragraph{Reverse attack.} Each of the previous attacks were designed to misclassify fake images as real. We find that it is also possible to generate adversarial perturbations that cause real images to be misclassified as fake. Somewhat surprisingly, this attack is harder, requiring a larger distortion: just under $7\%$ of the pixels must be flipped in a real image to lead to $50\%$ misclassification, as compared to $1\%$ of pixels required to lead to the same level of misclassification of a fake image (see Figure~\ref{fig:lpattacks}(b)).

% --------------------------------------------------------
\subsection{Loss-Maximizing Attack}

In this second attack, we define a simpler objective function that maximizes the likelihood that a fake image $x$ perturbed by $\delta$ is misclassified as real, but this time the $p$-norm of the distortion is fixed to be less than a specified threshold $\epsilon$. This optimization is formulated as:
\begin{eqnarray}
    \mathop{\text{arg min }}_{\delta \,\text{s.t.}\, \lVert \delta \rVert_p < \epsilon} f(x + \delta).
    \label{eqn:maxloss}
\end{eqnarray}
Unlike the previous Equation~(\ref{eqn:mindistortion}), this optimization is simpler because it does not require a search over the additional hyper-parameter. A standard gradient-descent optimization is used to solve for the optimal perturbation $\delta$~\cite{madry2018towards}.

\begin{figure}[t]
    \begin{tabular}{r@{\hspace{0.1cm}}r}
        (a) loss maximizing & (b) universal patch \\
        \includegraphics[width=0.48\linewidth]{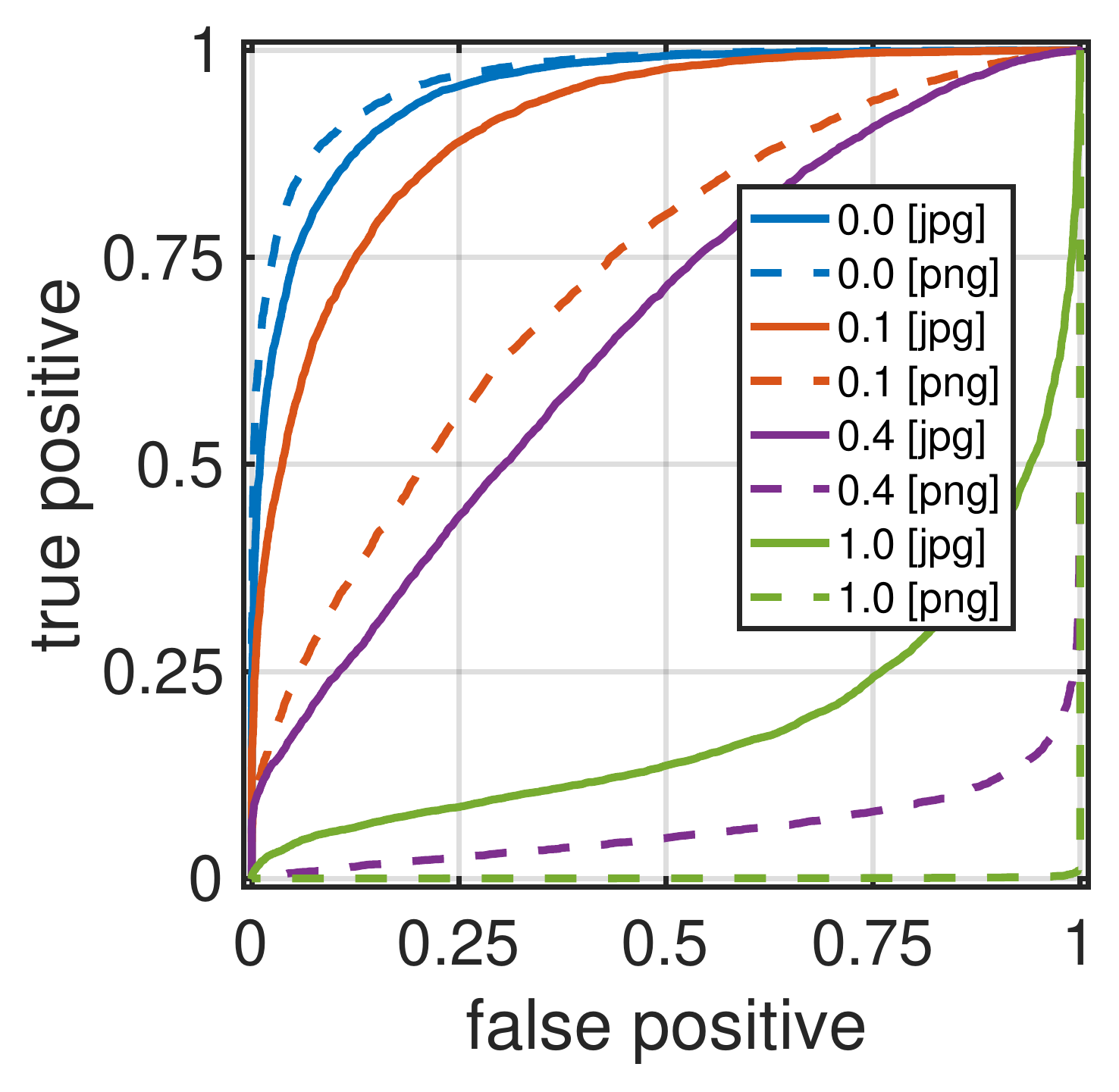} &
        \includegraphics[width=0.48\linewidth]{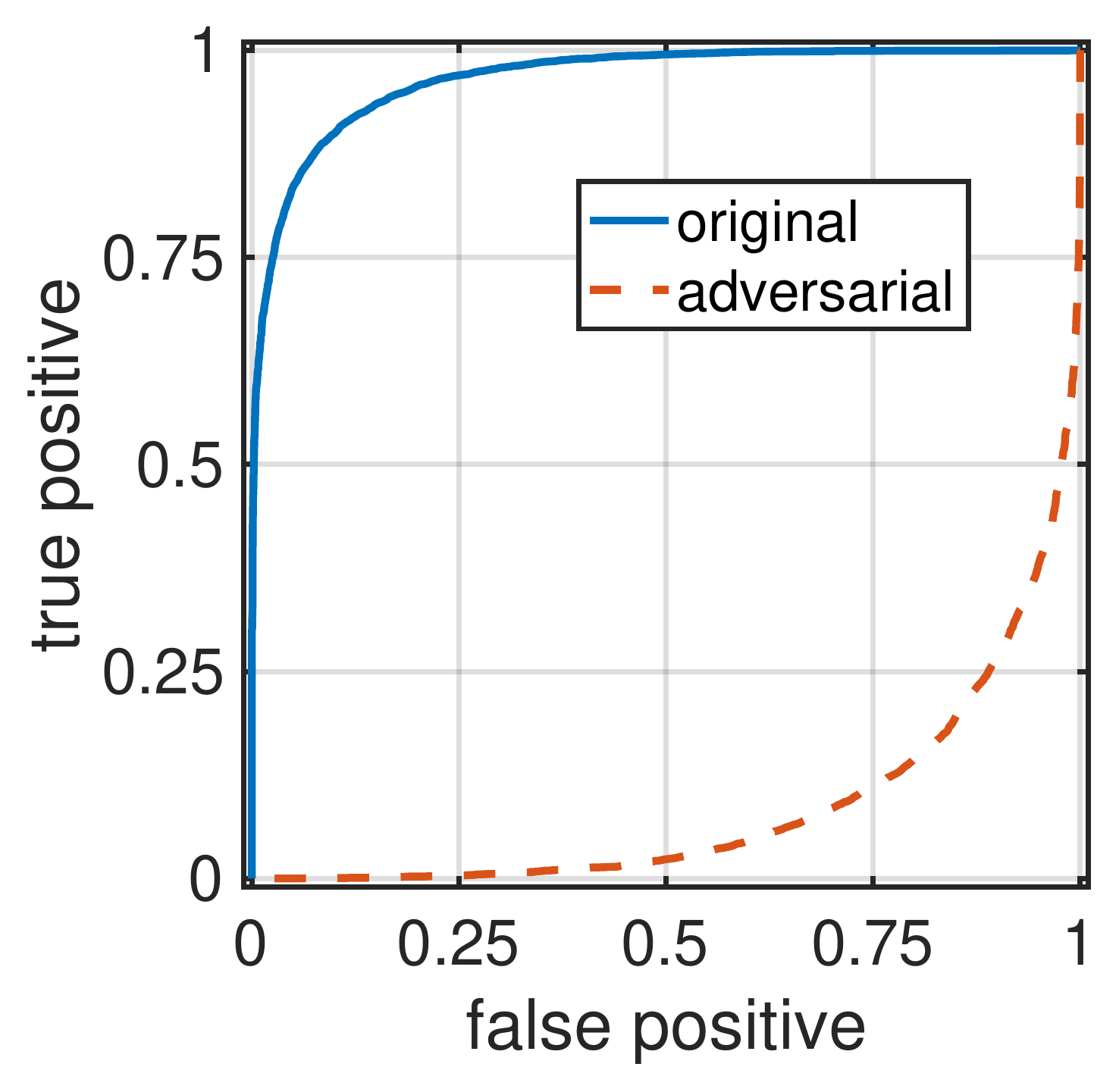} \\
        \\
        (c) latent space & (d) transfer \\
        \includegraphics[width=0.48\linewidth]{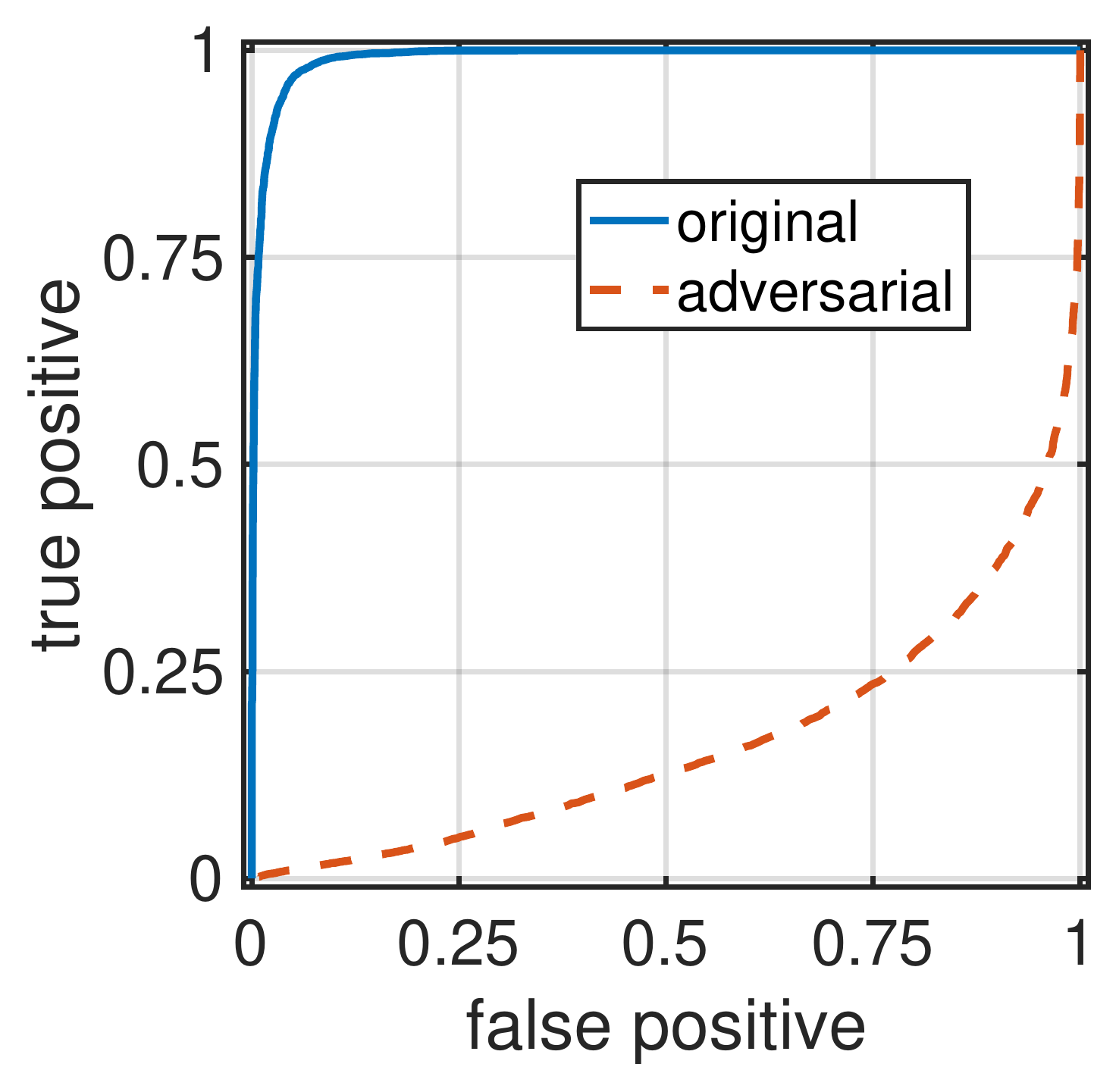} &
        \includegraphics[width=0.48\linewidth]{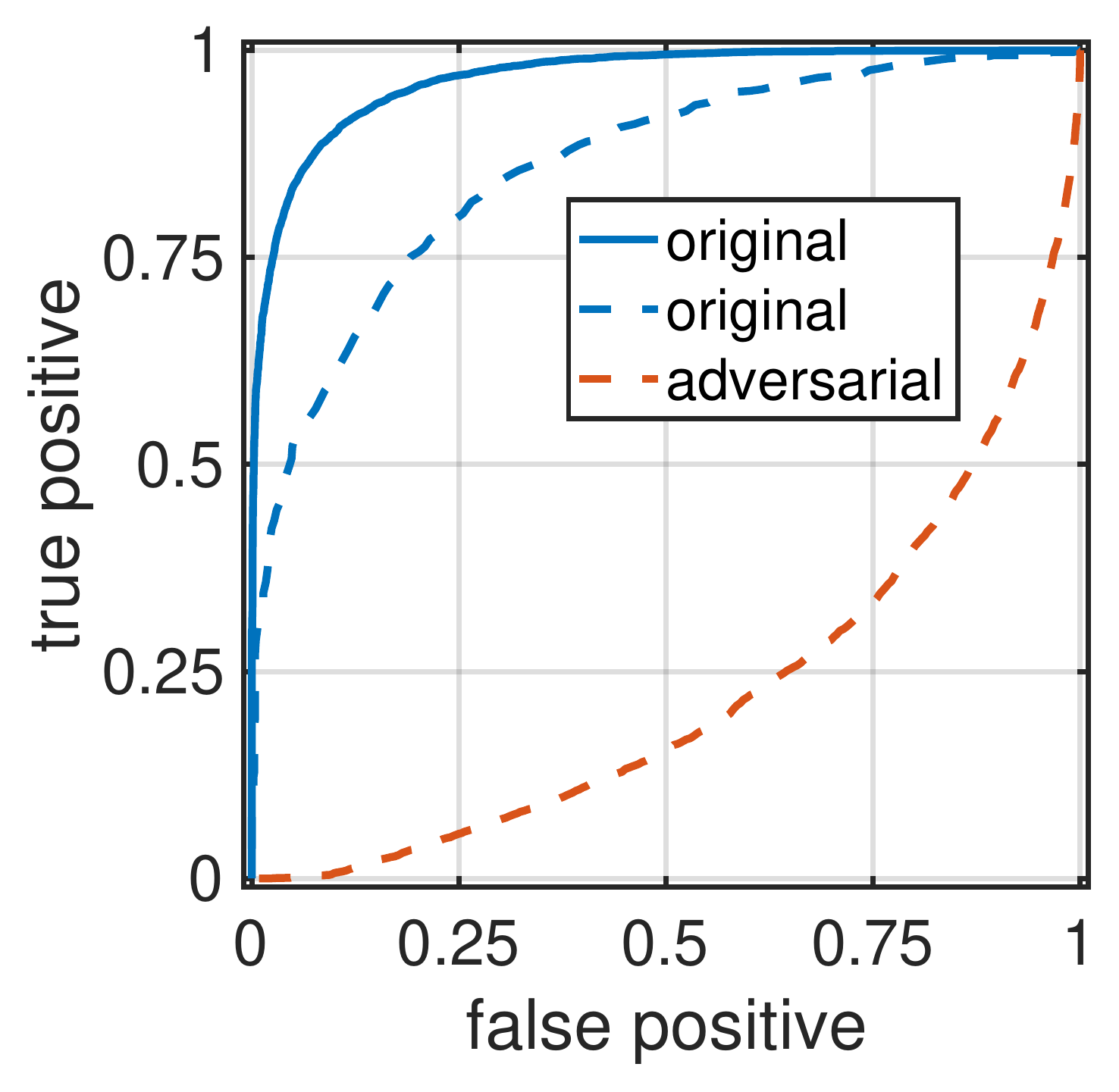}
    \end{tabular}
    \caption{Shown in each panel is the ROC curves for a forensic-classifier~\cite{wang20} before and after four distinct attacks: (a) classification accuracy for the originally synthesized images saved as JPEG (solid blue) and PNG (dashed blue) images and white-box adversarial images with varying fractions of flipped pixels; (b-c) classification accuracy for the originally synthesized  images (solid blue) and white-box adversarial images (dashed orange); and (d) classification accuracy for StyleGAN synthesized images for the forensic classifier of~\cite{wang20} (solid blue), our forensic classifier (dashed blue), and our black-box adversarial images (dashed orange).}
    \label{fig:ROC}
\end{figure}

This attack is also highly effective. Shown in Figure~\ref{fig:ROC}(a) is the trade-off between the false positive rate (incorrectly classifying a fake image as real) and the true positive rate (correctly classifying a fake image as fake) for a range of the fraction of modified pixels, between $0.0$ (non-adversarial) and $1.0$ (maximally adversarial). The solid curves correspond to the adversarial images saved in the JPEG format and the dashed curves correspond to the PNG format. Even with flipping the lowest-order bit of $40\%$ of pixels for uncompressed images, the AUC reduces from $0.966$ to $0.27$.

%
%

% --------------------------------------------------------
\subsection{Universal Adversarial-Patch Attack} 
\label{sec:universal-patch}

There is one significant limitation with the prior approaches in that the adversary is required to construct a tailored attack for each image -- at under $0.1$ seconds per image attack, our attacks are not especially costly, but the extra work may not be desirable.

To remedy this limitation, we create a single visible noise pattern that when overlaid on any fake image will result in the image being classified as real~\cite{brown2017adversarial}. Unlike the previous image-specific attacks, we generate a single universal patch that can be overlaid onto any fake image that then leads to misclassification. Similar to Equation~(\ref{eqn:maxloss}), the universal patch $\delta$ is generated by maximizing the expected loss of the classifier on a set of training examples $X$:
\begin{eqnarray}
    \mathop{\text{arg min }}_\delta \sum_{x \in X} \big[f(x_\delta))\big],
    \label{eqn:patch}
\end{eqnarray}
where $x_\delta$ denotes the input image $x$ overlaid with the patch $\delta$, fixed to be $1\%$ ($24 \times 24$ pixel) of the input image size.

A standard gradient-descent optimization is, again, used to maximize this objective function. On each gradient-descent iteration, a new image $x \in X$ is selected from a subset of $5,000$ images taken from the original $94,036$ image dataset, and disjoint from the $10,000$ evaluation images.

\begin{figure}
    \begin{tabular}{c@{\hspace{0.1cm}}c@{\hspace{0.1cm}}c}
        & adversarial fake & adversarial fake \\
        \raisebox{1.7cm}{(a)} &
        \includegraphics[width=0.44\linewidth]{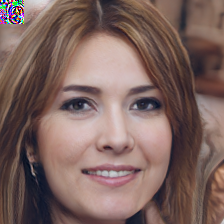} & 
        \includegraphics[width=0.44\linewidth]{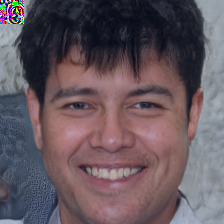} \\
        \\
        \hline \\
        & fake & adversarial fake \\
        \raisebox{1.7cm}{(b)} & 
        \includegraphics[width=0.44\linewidth]{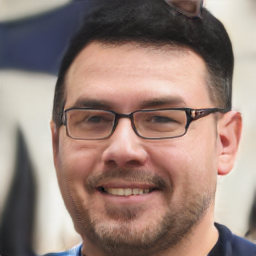} &
        \includegraphics[width=0.44\linewidth]{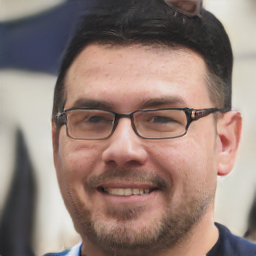} \\
        \raisebox{1.7cm}{(c)} & 
        \includegraphics[width=0.44\linewidth]{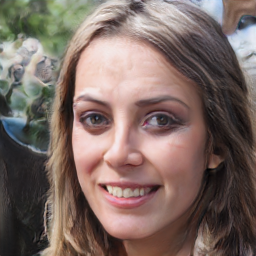} &
        \includegraphics[width=0.44\linewidth]{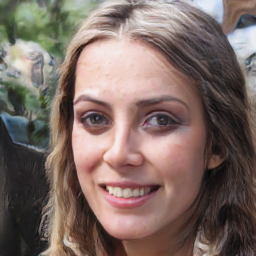} \\
        \raisebox{1.7cm}{(d)} & 
        \includegraphics[width=0.44\linewidth]{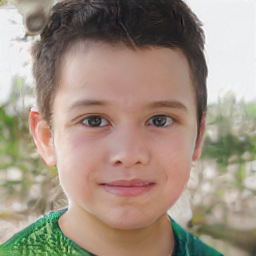} &
        \includegraphics[width=0.44\linewidth]{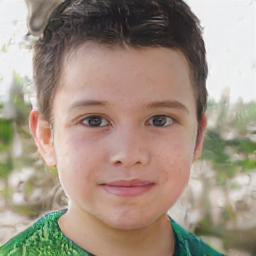}
    \end{tabular}
    \caption{Shown in row (a) are two adversarial examples in which a universal image patch is overlaid on a fake image causing it to be classified as real with high confidence. Shown in rows (b)-(d) are fake images (left) and their adversarial counterpart (right) created using a universal latent-space attack -- each of these adversarial images are misclassified as real with high confidence.}
    \label{fig:attack-examples}
\end{figure}

Shown in Figure~\ref{fig:attack-examples}(a) are two synthesized images with the overlaid patch (upper left corner) that are now classified as real with likelihood $98\%$ and $86\%$. Shown in Figure~\ref{fig:ROC}(b) is the trade-off between the false positive rate and the true positive rate for the  classifier when presented with the original images (solid blue curve) and the adversarial images (dashed orange curve). The AUC is reduced from $0.966$ to $0.085$.

%
%

% --------------------------------------------------------
\subsection{Universal Latent-Space Attack}
\label{sec:latent}

Each of the three previous attacks modified the input image $x$ by a perturbation $\delta$ to yield an adversarial image that is misclassified by the forensic classifier. In this fourth, and final, white-box attack, we introduce a latent-space attack on images in which the underlying representation used by the generative model is modified to yield an adversarial image. Here, we focus exclusively on images synthesized using StyleGAN~\cite{karras2018stylebased}.

Recall that we earlier formulated the generative model, $g \colon \mathcal{Z} \to \mathcal{X}$, as taking as input a vector in a latent space $z \in \mathcal{Z}$ and outputting a color image $x \in \mathcal{X}$. Recent generative models take two inputs, $g \colon \mathcal{Z} \times \mathcal{W} \to \mathcal{X}$, where $z \in \mathcal{Z}$ corresponds to high-level attributes and $w \in \mathcal{W}$ corresponds to low-level attributes. When synthesizing faces, for example, high-level attributes may correspond to gender, pose, skin color, and hair color or length, whereas low-level attributes may correspond to the presence of freckles. Our latent-space attack constructs a single (universal) attribute $\tilde{w} \in \mathcal{W}$ so that the resulting synthesized image, $g(z,\tilde{w})$ is misclassified by the forensic classifier $f(\cdot)$ as real.

As before, we apply a gradient-descent optimization to determine the universal adversarial perturbation. On each iteration, we sample a random latent vector $z$ and then maximize the loss of the classifier with respect to a single $\tilde{w}$. Specifically, we sample an initial random $\tilde{w}_0 \sim \mathcal{W}$ and then on each iteration $i$, update $\tilde{w}_{i+1} = \tilde{w}_i + \nabla_{\tilde{w}} f(g(z; \tilde{w}_i))$ where each $z \in \mathcal{Z}$ is chosen at random.

Shown in Figure~\ref{fig:attack-examples}(b)-(d) are representative examples of this attack. Shown in the left column are images synthesized with random, non-adversarial, attributes $w$. Shown in the right column are images synthesized with the universal adversarial attribute $\tilde{w}$. Although these images are not identical, they are nearly indistinguishable and the adversarial images are equally photorealistic.

Shown in Figure~\ref{fig:ROC}(c) is the trade-off between the false positive rate and the true positive rate for the classifier when presented with the original StyleGAN images (solid blue curve) and the adversarial images (dashed orange curve). The AUC is reduced from $0.99$ to $0.17$.

% ------------------------------------------------------
\section{Black-Box Attack}
\label{sec:black-box}

In practice, forensic classifiers for detecting synthesized images may not be publicly released in order to make it difficult to launch the type of white-box attacks we describe previously. Social networks or dating sites, for example, might analyze all portrait photos and reject or label synthetically-generated images. In such cases, an adversary will not be able to perform gradient-descent optimizations on the model weights described in the previous section.

A \emph{black-box} threat model corresponds to the situation when the adversary does not have access to the exact details of the forensic classifier, but is aware what type of classifier is in place. Defenses are only interesting if they remain secure against an adversary who is aware they are present; ``security through obscurity'' is not a valid defense~\cite{anderson2008security}. As such, the black-box threat model assumes that the adversary knows that there is a defense in place, and similarly knows the general strategy of the defense.

We find that forensic classifiers are vulnerable to adversarial examples even in this more restrictive black-box setting. To mount these attacks, we make use of the transferability property of adversarial examples~\cite{papernot2016transferability,liu2016delving}. Consider two forensic classifiers, a source classifier $f_1(\cdot)$ and a target classifier $f_2(\cdot)$, both trained to perform the same task. An adversarial input $x$ that is misclassified by $f_1(\cdot)$ is also likely to be misclassified by $f_2(\cdot)$, even if these source and target classifiers are different neural-network architectures, trained with different algorithms on different training data. This source to target transferability works because the decision boundaries learned by different classifiers, solving the same underlying problem, are sufficiently similar that adversarial examples transfer between classifiers~\cite{liu2016delving}.

In order to transfer an attack to the target detector of Wang \emph{et al.}~\cite{wang20}, we develop our own forensic classifier. We collect one million ProGAN-generated images and one million real images on which ProGAN was trained. We then train a ResNet-18 (as compared to Wang \emph{et al.}'s ResNet-50) to classify images as fake or real. Unlike Wang \emph{et al.}, our classifier was not pre-trained on ImageNet. We apply the standard ImageNet transformations: randomly crop a $224 \times 224$ pixel bounding box from each input image; and horizontally flip each real image with a $50\%$ probability.

Our classifier achieves an overall AUC of $0.85$, significantly less than the AUC of $0.96$ achieved by Wang \emph{et al.}. As we will see next, despite the differences in the training arhitecture and data, and the lower accuracy of our source classifier, we are still able to transfer an attack from our classifier to Wang \emph{et al.}'s classifier.

We first generated white-box adversarial examples on our source classifier using the previously described $\ell_0$-distortion minimizing attack (with a distortion bound of $1/255$ per pixel -- the smallest distortion bound that can still be saved to a PNG). These adversarial examples reduce our source classifier's AUC from $0.85$ to $0.003$. Shown in Figure~\ref{fig:ROC}(d) is the effect of transferring these exact adversarial examples to Wang \emph{et al.}'s classifier. This attack reduces the target classifier AUC from $0.96$ to $0.22$.

While this is the weakest of all attacks that we have presented it is not weak as it drives the forensic classifier to below chance performance. This attack is also the most realistic since in practice, attacks will likely need to be mounted within this black-box threat model.

% --------------------------------------------------------
\section{Discussion}

\paragraph{Motivating the Rules.} Having demonstrated the vulnerability of forensic classifiers under the standard perturbation-based, adversarial machine-learning threat model, we now turn to the fundamental question posed by Gilmer \emph{et al.}~\cite{gilmer2018motivating}: is an actual adversary limited to perturbation attacks? We believe that the answer in this setting is firmly \emph{no}. It is not realistic to require that an adversary only apply an indistinguishable perturbation to an image to cause misclassification. True adversaries will have a much larger space of valid actions to operate under. For example, even standard image laundering -- resizing, rescaling, cropping, or recompression -- often reduces the true positive rate by over ten percentage points. A naive adversary might still succeed through these techniques alone, without needing to resort to more powerful, but also more complicated, attacks.

Further, an adversary does not necessarily need one particular image to be identified as real, but rather some semantically similar image to be classified as real. For example, the exact image of Mr. Walz shown in Figure~\ref{fig:andrew-walz} was not essential to create a fictional congressional candidate's Twitter account -- any number of photorealistic portrait photos would have sufficed. As such, even if Twitter was using a forensic classifier to scan portrait photos for synthetically-generated content, an adversary would need only repeatedly upload different photos until one simply failed detection. Even with a relatively high true positive rate of $90\%$, an adversary would need only upload, on average, ten images before the classifier failed to detect a fake image.

Even though we only considered attacks that are harder than the attacks that might actually be applied in reality, we still believe that it is worthwhile to study this worst-case, low-distortion perturbation attacks. While clearly this is not the only possible attack, it is highly restrictive and therefore difficult to execute. Given the relative ease with which we were able to make this restrictive attack succeed, other attacks with fewer constraints are likely to be even easier to execute.

\smallskip
\paragraph{Who goes first?}
A second important question to consider in these types of defender/forger situations is which agent goes first and which agent has to react. In a traditional situation, the defender acts first, releasing some product or service, and then the forger responds, looking for vulnerabilities. In this situation, the forger has the advantage because she need only construct one successful attack whereas the defender has to prevent all possible attacks. In other scenarios, the forger commits to a particular approach and the defender reacts, adjusting her defenses accordingly. In practice, either scenario is possible. Twitter might, for example, deploy a forensic classifier to classify uploaded profile photos as real or fake. The forger could then modify her generator to defeat the classifier. Alternatively, a fact-checking organization might retroactively apply a forensic classifier over historical news photos. In this scenario, the defender is likely to know the forger's potential synthesis techniques.

Ultimately, the ordering is effectively a matter of the time-scale being considered. On a relatively short time-scale of hours to days, if the forger goes second, then she will have the advantage. On a longer time-scale of months to years, the defender will eventually have knowledge of the forger's techniques and will have the advantage of retroactively finding the fakes. On the internet, however, where, for example, the half-life of a tweet is on the order of minutes, the game is over in the first few hours, giving the forger an inherent advantage. 

\begin{figure}[t]
    \centering
    \includegraphics[width=0.48\linewidth]{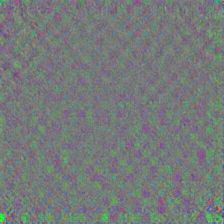}
    \includegraphics[width=0.48\linewidth]{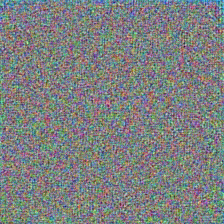}
    \caption{Mean perturbation for the forensics classifier of~\cite{wang20} (left) and an ImageNet classifier (right) needed to lead to misclassification.}
    \label{fig:mean}
\end{figure}

\smallskip
\paragraph{Classifier Sensitivity.} We find that the forensic detectors require perturbations roughly $10$ times smaller than necessary to fool ImageNet classifiers, also operating on $224 \times 224$ images. In order to better understand why these detectors are so sensitive, we compute the average perturbation necessary to fool the forensic classifier of~\cite{wang20}. This is done by averaging the adversarial perturbation introduced on $2000$ adversarial StyleGAN examples in the released dataset of~\cite{wang20}. Shown in Figure~\ref{fig:mean} is a contrast-enhanced version of this average perturbation and, for comparison, the average perturbation required to fool an ImageNet ResNet-50 classifier (the same architecture used by the forensic classifier). The forensic-classifier perturbation consists of highly reqular structure aligned with the $8 \times 8$ JPEG lattice. We suspect, but have not yet confirmed, that this points to a possible JPEG artifact in the underlying training data.

\smallskip
\paragraph{Counter-Defenses.} Extensive efforts have attempted to defend against adversarial examples on standard image classification~\cite{metzen2017detecting,xu2017feature,meng2017magnet,buckman2018thermometer,xiao2019resisting}. Almost all proposed defenses, however, have been shown to be ineffective at increasing classification robustness, and leave accuracy at $0\%$ even under small distortion bounds~\cite{athalye2018obfuscated,tramer2020adaptive}. The two most effective defenses on large images have been adversarial training~\cite{madry2018towards} and randomized smoothing~\cite{lecuyer2019certified,cohen2019certified}. Adversarial training continuously trains on adversarial examples generated on previous versions of the classifier. In contrast, randomized smoothing adds large magnitude, Gaussian noise to every pixel, (provably) making it impossible for any small perturbation to change the classifier output. We believe that it would be interesting to study the efficacy of these two counter-defense strategies on detecting synthesized images. Because adversarial training only offers limited robustness on traditional image classification tasks, and because detecting synthesized images is an even harder problem, it may be difficult to obtain meaningful robustness through adversarial training alone.

% ------------------------------------------------------
\section{Conclusions}
\label{sec:conclusion}

To the extent that synthesized or manipulated content is used for nefarious purposes, the problem of detecting this content is inherently adversarial. We argue, therefore, that forensic classifiers need to build an adversarial model into their defenses. This model must go beyond the standard laundering attacks of recompression, resizing, blurring, or adding white noise. 

Adversarial modeling is particularly important in the types of data-driven, machine-learning based techniques described here. We have shown that these techniques, are highly vulnerable to attack because the same power and flexibility of the underlying neural-network classifiers that leads to high classification accuracies, can also be easily manipulated to create adversarial images that easily subvert detection. This subversion takes the form of white-box attacks in which it is assumed that the details of the forensic classifier are known, and black-box attacks in which it is assumed that only a forensic classifier, of unknown detail, exists. These attacks can imperceptibly modify fake images so that they are misclassified as real, or imperceptibly modify real images so that they are misclassified as fake.

It may be argued that white-box attacks are not a significant threat because, in critical scenarios, the details of the forensic classifier can be withheld. We have shown, however, the efficacy of black-box attacks in which the classifier details are not known -- the threat posted by these attacks is surely more significant.

We have shown the efficacy of these types of attacks on two previously published forensic classifiers, and a classifier of our own creation. Previous results from the adversarial machine learning literature~\cite{szegedy13,carlini2017towards}, however, suggest that this vulnerability is inherent to all neural-network based forensic classifiers \cite{ilyas2019adversarial}.

Demonstrating attacks on sensitive systems is not something that should be taken lightly, or done simply for sport.  However, if such forensic classifiers are currently deployed, the false sense of security they provide may be worse than if they were not deployed at all -- not only would a fake profile picture appear authentic, now it would be given additional credibility by a forensic classifier.

Even if forensic classifiers are eventually defeated by a committed adversary, these classifiers are still valuable in that they make it more difficult and time consuming to create a convincing fake. They would, for example, have made it more difficult for a 17-year old high school student to create a realistic image to use in the creation of a fictional congressional candidate on social media. While this is unarguably a low bar, continued efforts to increase the resilience of forensic classifiers will raise this bar, eventually  making it more difficult for the average person to distribute convincing and undetectable deep-fake images.

\newpage

%\iffalse
\section*{Acknowldgements}
We thank Joel Frank, Andrew Owens, Alexei Efros, and Sheng Yu Wang for helpful discussions and assistance with running their detectors. We additionally thank David Berthelot, Andreas Terzis, and Carey Radebaugh for feedback on earlier drafts of this paper. This research was developed with funding from the Defense Advanced Research Projects Agency (DARPA FA8750-16-C-0166). The views, opinions, and findings expressed are those of the authors and should not be interpreted as representing the official views or policies of the Department of Defense or the U.S. Government.
%\fi

% ------------------------------------------------------

{\small
\bibliographystyle{ieee_fullname}
\bibliography{main}

\begin{thebibliography}{10}\itemsep=-1pt

\bibitem{katiejones19}
Experts: Spy used {AI}-generated face to connect with targets.
\newblock \url{https://apnews.com/bc2f19097a4c4fffaa00de6770b8a60d}.
\newblock Accessed: 2020-03-12.

\bibitem{andrewwalz20}
A high school student created a fake 2020 us candidate. {T}witter verified it.
\newblock
  \url{https://www.cnn.com/2020/02/28/tech/fake-twitter-candidate-2020/index.html}.
\newblock Accessed: 2020-03-12.

\bibitem{fakefaces20}
How fake faces are being weaponized online.
\newblock
  \url{https://www.cnn.com/2020/02/20/tech/fake-faces-deepfake/index.html}.
\newblock Accessed: 2020-03-12.

\bibitem{agarwal2019}
Shruti Agarwal, Hany Farid, Yuming Gu, Mingming He, Koki Nagano, and Hao Li.
\newblock Protecting world leaders against deep fakes.
\newblock In {\em IEEE Conference on Computer Vision and Pattern Recognition,
  Workshop on Media Forensics}, pages 38--45, 2019.

\bibitem{anderson2008security}
Ross Anderson.
\newblock {\em Security Engineering}.
\newblock John Wiley \& Sons, 2008.

\bibitem{athalye2018obfuscated}
Anish Athalye, Nicholas Carlini, and David Wagner.
\newblock {Obfuscated gradients give a false sense of security: Circumventing
  defenses to adversarial examples}.
\newblock {\em arXiv: 1802.00420}, 2018.

\bibitem{bohme13}
Rainer B{\"o}hme and Matthias Kirchner.
\newblock {Counter-forensics: Attacking image forensics}.
\newblock In {\em Digital image forensics}, pages 327--366. Springer, 2013.

\bibitem{biggan18}
Andrew Brock, Jeff Donahue, and Karen Simonyan.
\newblock {Large scale GAN training for high fidelity natural image synthesis}.
\newblock arXiv: 1809.11096, 2018.

\bibitem{brown2017adversarial}
Tom~B Brown, Dandelion Man{\'e}, Mart{\'\i}n~Abadi Aurko~Roy, and Justin
  Gilmer.
\newblock Adversarial patch.
\newblock 2017.

\bibitem{buckman2018thermometer}
Jacob Buckman, Aurko Roy, Colin Raffel, and Ian Goodfellow.
\newblock Thermometer encoding: One hot way to resist adversarial examples.
\newblock In {\em International Conference on Learning Representations}, 2018.

\bibitem{carlini2017towards}
Nicholas Carlini and David Wagner.
\newblock Towards evaluating the robustness of neural networks.
\newblock In {\em IEEE Symposium on Security and Privacy}, pages 39--57, 2017.

\bibitem{carlini18}
Nicholas Carlini and David Wagner.
\newblock {Audio adversarial examples: Targeted attacks on speech-to-text}.
\newblock In {\em IEEE Security and Privacy Workshops}, pages 1--7, 2018.

\bibitem{stargan18}
Yunjey Choi, Minje Choi, Munyoung Kim, Jung-Woo Ha, Sunghun Kim, and Jaegul
  Choo.
\newblock {Stargan: Unified generative adversarial networks for multi-domain
  image-to-image translation}.
\newblock In {\em IEEE International Conference on Computer Vision}, pages
  8789--8797, 2018.

\bibitem{ciftci2019}
Umur~Aybars Ciftci and Ilke Demir.
\newblock Fakecatcher: Detection of synthetic portrait videos using biological
  signals.
\newblock arXiv: 1901.02212, 2019.

\bibitem{cohen2019certified}
Jeremy~M Cohen, Elan Rosenfeld, and J~Zico Kolter.
\newblock Certified adversarial robustness via randomized smoothing.
\newblock {\em arXiv:1 902.02918}, 2019.

\bibitem{cozzolino18}
Davide Cozzolino, Justus Thies, Andreas R{\"o}ssler, Christian Riess, Matthias
  Nie{\ss}ner, and Luisa Verdoliva.
\newblock {Forensictransfer: Weakly-supervised domain adaptation for forgery
  detection}.
\newblock {\em arXiv: 1812.02510}, 2018.

\bibitem{deng09}
Jia Deng, Wei Dong, Richard Socher, Li-Jia Li, Kai Li, and Li Fei-Fei.
\newblock {Imagenet: A large-scale hierarchical image database}.
\newblock In {\em IEEE Conference on Computer Vision and Pattern Recognition},
  pages 248--255, 2009.

\bibitem{ebrahimi2017hotflip}
Javid Ebrahimi, Anyi Rao, Daniel Lowd, and Dejing Dou.
\newblock {HotFlip: White-Box Adversarial Examples for Text Classification}.
\newblock 2017.

\bibitem{frank2020leveraging}
Joel Frank, Thorsten Eisenhofer, Lea Schönherr, Asja Fischer, Dorothea
  Kolossa, and Thorsten Holz.
\newblock Leveraging frequency analysis for deep fake image recognition.
\newblock 2020.

\bibitem{g2020adversarial}
Apurva Gandhi and Shomik Jain.
\newblock Adversarial perturbations fool deepfake detectors.
\newblock arXiv: 2003.10596, 2020.

\bibitem{gilmer2018motivating}
Justin Gilmer, Ryan~P Adams, Ian Goodfellow, David Andersen, and George~E Dahl.
\newblock Motivating the rules of the game for adversarial example research.
\newblock 2018.

\bibitem{he16}
Kaiming He, Xiangyu Zhang, Shaoqing Ren, and Jian Sun.
\newblock {Deep residual learning for image recognition}.
\newblock In {\em IEEE Conference on Computer Vision and Pattern Recognition},
  pages 770--778, 2016.

\bibitem{ilyas2019adversarial}
Andrew Ilyas, Shibani Santurkar, Dimitris Tsipras, Logan Engstrom, Brandon
  Tran, and Aleksander Madry.
\newblock Adversarial examples are not bugs, they are features.
\newblock In {\em Advances in Neural Information Processing Systems}, pages
  125--136, 2019.

\bibitem{karras2017progressive}
Tero Karras, Timo Aila, Samuli Laine, and Jaakko Lehtinen.
\newblock {Progressive Growing of GANs for Improved Quality, Stability, and
  Variation}.
\newblock arXiv: 1710.10196, 2017.

\bibitem{karras2018stylebased}
Tero Karras, Samuli Laine, and Timo Aila.
\newblock {A Style-Based Generator Architecture for Generative Adversarial
  Networks}.
\newblock arXiv: 1812.04948, 2018.

\bibitem{karras2019analyzing}
Tero Karras, Samuli Laine, Miika Aittala, Janne Hellsten, Jaakko Lehtinen, and
  Timo Aila.
\newblock {Analyzing and Improving the Image Quality of StyleGAN}.
\newblock arXiv: 1912.04958, 2019.

\bibitem{kingma2014adam}
Diederik~P Kingma and Jimmy Ba.
\newblock {Adam: A method for stochastic optimization}.
\newblock {\em arXiv: 1412.6980}, 2014.

\bibitem{lecuyer2019certified}
Mathias Lecuyer, Vaggelis Atlidakis, Roxana Geambasu, Daniel Hsu, and Suman
  Jana.
\newblock Certified robustness to adversarial examples with differential
  privacy.
\newblock In {\em 2019 IEEE Symposium on Security and Privacy}, pages 656--672,
  2019.

\bibitem{li2018blinking}
Yuezun Li, Ming-Ching Chang, and Siwei Lyu.
\newblock In ictu oculi: Exposing {AI} created fake videos by detecting eye
  blinking.
\newblock In {\em IEEE International Workshop on Information Forensics and
  Security}, pages 1--7, 2018.

\bibitem{liu2016delving}
Yanpei Liu, Xinyun Chen, Chang Liu, and Dawn Song.
\newblock Delving into transferable adversarial examples and black-box attacks.
\newblock 2016.

\bibitem{madry2018towards}
Aleksander Madry, Aleksandar Makelov, Ludwig Schmidt, Dimitris Tsipras, and
  Adrian Vladu.
\newblock {Towards Deep Learning Models Resistant to Adversarial Attacks}.
\newblock {\em International Conference on Learning Representations}, 2018.

\bibitem{marra18}
Francesco Marra, Diego Gragnaniello, Davide Cozzolino, and Luisa Verdoliva.
\newblock {Detection of GAN-generated fake images over social networks}.
\newblock In {\em IEEE Conference on Multimedia Information Processing and
  Retrieval}, pages 384--389, 2018.

\bibitem{meng2017magnet}
Dongyu Meng and Hao Chen.
\newblock Magnet: a two-pronged defense against adversarial examples.
\newblock In {\em ACM SIGSAC Conference on Computer and Communications
  Security}, pages 135--147, 2017.

\bibitem{metzen2017detecting}
Jan~Hendrik Metzen, Tim Genewein, Volker Fischer, and Bastian Bischoff.
\newblock On detecting adversarial perturbations.
\newblock 2017.

\bibitem{neekhara20}
Paarth Neekhara, Shehzeen Hussain, Malhar Jere, Farinaz Koushanfar, and Julian
  McAuley.
\newblock {Adversarial Deepfakes: Evaluating Vulnerability of Deepfake
  Detectors to Adversarial Examples}.
\newblock 2020.

\bibitem{papernot2016transferability}
Nicolas Papernot, Patrick McDaniel, and Ian Goodfellow.
\newblock Transferability in machine learning: from phenomena to black-box
  attacks using adversarial samples.
\newblock 2016.

\bibitem{gaugan19}
Taesung Park, Ming-Yu Liu, Ting-Chun Wang, and Jun-Yan Zhu.
\newblock {Semantic image synthesis with spatially-adaptive normalization}.
\newblock In {\em IEEE Conference on Computer Vision and Pattern Recognition},
  pages 2337--2346, 2019.

\bibitem{rossler19}
Andreas Rossler, Davide Cozzolino, Luisa Verdoliva, Christian Riess, Justus
  Thies, and Matthias Nie{\ss}ner.
\newblock {Faceforensics++: Learning to detect manipulated facial images}.
\newblock In {\em IEEE International Conference on Computer Vision}, pages
  1--11, 2019.

\bibitem{ruiz2020disrupting}
Nataniel Ruiz, Sarah~Adel Bargal, and Stan Sclaroff.
\newblock Disrupting deepfakes: Adversarial attacks against conditional image
  translation networks and facial manipulation systems.
\newblock 2020.

\bibitem{szegedy13}
Christian Szegedy, Wojciech Zaremba, Ilya Sutskever, Joan Bruna, Dumitru Erhan,
  Ian Goodfellow, and Rob Fergus.
\newblock {Intriguing properties of neural networks}.
\newblock 2013.

\bibitem{tramer2020adaptive}
Florian Tramer, Nicholas Carlini, Wieland Brendel, and Aleksander Madry.
\newblock On adaptive attacks to adversarial example defenses.
\newblock {\em arXiv: 2002.08347}, 2020.

\bibitem{wang20}
Sheng-Yu Wang, Oliver Wang, Richard Zhang, Andrew Owens, and Alexei~A Efros.
\newblock {CNN-generated images are surprisingly easy to spot...for now}.
\newblock In {\em IEEE Conference on Computer Vision and Pattern Recognition},
  2020.

\bibitem{xiao2019resisting}
Chang Xiao, Peilin Zhong, and Changxi Zheng.
\newblock Resisting adversarial attacks by $k$-winners-take-all.
\newblock 2019.

\bibitem{xu2017feature}
Weilin Xu, David Evans, and Yanjun Qi.
\newblock Feature squeezing: Detecting adversarial examples in deep neural
  networks.
\newblock 2017.

\bibitem{yang2019}
Xin Yang, Yuezun Li, and Siwei Lyu.
\newblock Exposing deep fakes using inconsistent head poses.
\newblock In {\em IEEE International Conference on Acoustics, Speech and Signal
  Processing}, pages 8261--8265, 2019.

\bibitem{yu18}
Ning Yu, Larry Davis, and Mario Fritz.
\newblock {Attributing fake images to GANs: Analyzing fingerprints in generated
  images}.
\newblock {\em arXiv:1811.08180}, 2018.

\bibitem{zhang19}
Xu Zhang, Svebor Karaman, and Shih-Fu Chang.
\newblock {Detecting and simulating artifacts in GAN fake images}.
\newblock {\em arXiv: 1907.06515}, 2019.

\bibitem{cyclegan17}
Jun-Yan Zhu, Taesung Park, Phillip Isola, and Alexei~A Efros.
\newblock {Unpaired image-to-image translation using cycle-consistent
  adversarial networks}.
\newblock In {\em IEEE International Conference on Computer Vision}, pages
  2223--2232, 2017.

\end{thebibliography}
}

% ------------------------------------------------------

\end{document}